\documentclass[pdflatex,sn-apa]{sn-jnl}
\usepackage{graphicx}%
\usepackage{multirow}%
\usepackage{amsmath,amssymb,amsfonts}%
\usepackage{amsthm}%
\usepackage{mathrsfs}%
\usepackage[title]{appendix}%
\usepackage{xcolor}%
\usepackage{textcomp}%
\usepackage{manyfoot}%
\usepackage{booktabs}%
\usepackage{algorithm}%
\usepackage{algorithmicx}%
\usepackage{algpseudocode}%
\usepackage{listings}%
\usepackage{fvextra}
\theoremstyle{thmstyleone}%
\theoremstyle{thmstyletwo}%
\theoremstyle{thmstylethree}%
\begin{document}

\title[Patterns in Medical Item Difficulty]{
What Makes Medical Examination Items Difficult?
Using Large Language Models to Explore Patterns
Associated With Empirical Item Difficulty
}

\author*[1]{\fnm{Xiaomeng} \sur{Xiong}}
\email{xiaomengxiong@ufl.edu}

\author[2]{\fnm{Marcus} \sur{Walker}}

\author[2]{\fnm{Yanlin} \sur{Jiang}}

\affil*[1]{\orgname{University of Florida}, \orgaddress{\city{Gainesville}, \state{Florida}, \country{USA}}}

\affil[2]{\orgname{National Commission on Certification of Physician Assistants}, \orgaddress{\country{USA}}}

\abstract{
Research on medical item difficulty has largely focused on prediction rather than on identifying interpretable patterns associated with empirical difficulty. This study used empirical difficulty labels to guide large language models (LLMs) in inducing difficulty rubrics and examined how rubric guidance affected LLM difficulty judgments.

Using training items and their empirical difficulty estimates from the United States Medical Licensing Examination (USMLE) and the Physician Assistant National Certifying Examination (PANCE), Claude Opus 5 induced separate rubrics for absolute and pairwise difficulty judgment. None of the induced rubrics treated surface features, such as stem length, option count, or images, as major sources of difficulty. Several patterns emerged across both examinations, including cue convergence, misleading cues, and option competition, although their specific forms differed. Retrieval of knowledge that could not be derived from the stem was particularly prominent in the PANCE rubrics, whereas the USMLE rubrics placed greater emphasis on integrating multiple clinical cues.

We compared commercial and open-weight LLMs on held-out items under unguided and rubric-guided conditions. Rubric guidance changed how models used the difficulty scale but did not consistently improve performance. In absolute judgments, guidance encouraged broader use of the scale and generally improved coarse-grained agreement, without consistently improving rank ordering. In pairwise comparisons, its effects varied across models and examinations, and judgments became more sensitive to presentation order. These findings suggest that the value of label-guided rubric induction may lie more in generating interpretable, testable hypotheses from empirical difficulty data than in reliably improving difficulty judgments.}

\keywords{Item difficulty, Medical licensure examinations, Large language models, Rubric induction, Interpretability, Educational assessment}

\maketitle

\section{Introduction}\label{sec1}

Item difficulty is a central consideration in test development and item bank maintenance. A substantial body of research has attempted to predict empirical item difficulty using item text, manually constructed item features, language representations, or simulated examinee responses \citep{duenas2024upn,gombert2024predicting,ha-etal-2019-predicting,park2024large,yaneva2024findings}. Although these approaches address whether difficulty can be predicted, predictive performance alone does not explain why some items are more difficult than others. A representation may capture signals associated with difficulty without yielding principles that item writers can readily inspect, interpret, and apply.

Large language models offer a different way to approach this problem. Beyond directly predicting item difficulty, LLMs can synthesize information across many item examples, compare recurring patterns, and summarize those patterns in natural language. The present study therefore focuses not only on using LLMs to predict difficulty, but on using empirical difficulty information to guide the induction of interpretable principles that distinguish easier from harder items. In this framework, empirical difficulty is not only an outcome to be predicted. It can also serve as supervision for identifying recurring structures that are systematically associated with item difficulty.

We used two medical licensure examination datasets, the United States Medical Licensing Examination (USMLE) and the Physician Assistant National Certifying Examination (PANCE). A frontier LLM, Claude Opus 5, was used to induce separate rubrics for absolute difficulty judgment and pairwise difficulty comparison from training items and their empirical difficulty. We then compared unguided and rubric-guided judgments of new items across multiple commercial and open-weight models. This design allowed us to examine whether empirical difficulty labels could be translated into an interpretable account of medical item difficulty, how the identified patterns differed across the two examinations, and how making those patterns explicit affected LLM judgments under absolute and pairwise formats. Accordingly, the study addressed the following research questions:

RQ1. What item difficulty patterns are identified through LLM rubric induction, and how do these patterns differ across the two medical licensure examinations?

RQ2. How does rubric guidance affect LLM judgments of medical item difficulty across absolute and pairwise judgment formats?

\section{Related Work}\label{sec2}

\subsection{Medical Item Difficulty Prediction}
Prior work on medical item difficulty prediction has examined whether empirical difficulty can be estimated from item content and other available item information. Early work used manually constructed linguistic features, word embeddings, and information retrieval features to predict the difficulty of USMLE items \citep{ha-etal-2019-predicting}. The BEA 2024 shared task later provided a common public benchmark and included approaches based on fine-tuned transformer models, general-purpose and biomedical text representations, and other features derived from item content \citep{gombert2024predicting,yaneva2024findings}. Other work has taken a different approach by using LLMs to simulate examinee responses and estimating item difficulty from simulated performance \citep{duenas2024upn,park2024large}.

Despite the different signals used by these approaches, existing results suggest that empirical difficulty is difficult to recover from the items themselves. \citet{ha-etal-2019-predicting} combined word embeddings, linguistic features, and information retrieval features across 12,038 USMLE Clinical Knowledge items, yet their full model achieved a correlation of only $r = .32$ with empirical difficulty. The strongest correlation for any individual feature was approximately .11. Performance in the BEA 2024 shared task was also limited. The best-performing system achieved a Pearson correlation of .27, a Spearman correlation of .25, and an RMSE of .299, only a modest improvement over a mean-prediction baseline with an RMSE of .311 \citep{gombert2024predicting,yaneva2024findings}. These findings suggest that even when different types of item information and modeling approaches are combined, empirical difficulty remains difficult to predict accurately from the item alone.

More importantly, whether these approaches rely on manually constructed features, text representations, or simulated responses, their primary goal is to predict empirical difficulty rather than to explain the structures that give rise to it. Comparatively little work has focused on deriving readable and comparable accounts of item difficulty directly from empirical difficulty itself.

\subsection{Interpretable Modeling of Item Difficulty}
\citet{razavi2026estimating} decomposed the overall task of judging item difficulty by asking an LLM to evaluate each item on a predefined set of features and then using those feature ratings as inputs to tree-based models. They defined 20 features for mathematics items and 13 for reading items. These features were developed through expert focus groups and cross-checked against the existing literature. Across 5,170 K--5 mathematics and reading items, this approach achieved correlations as high as $r = .87$.

However, this approach depends on a predefined feature set. For medical items, no comparable framework is currently available. Prior work has shown that surface complexity and semantic representations are relatively weak predictors of medical item difficulty, but there is no established set of features that can be used to explain empirical difficulty in this domain.

There is precedent for constructing such feature sets automatically. The Text Bottleneck Model \citep[TBM;][]{ludan2023interpretable} extends the Concept Bottleneck Model \citep[CBM;][]{koh2020concept} from computer vision to text and uses two LLM modules for concept generation and concept measurement. This reduces the need for researchers to manually construct a concept list or annotate concept values for every example.

In TBM, however, candidate concepts are proposed primarily from the text itself, while labels are used later to determine whether a concept should be retained based on whether it improves predictive performance. This mechanism is well suited to identifying attributes that can be directly recognized and described in the text. For example, in restaurant reviews, TBM identifies concepts such as service quality, food quality, and price perception, all of which are explicitly reflected in the review content \citep{ludan2023interpretable}. Medical item difficulty may have a different structure. Findings from prior studies and the BEA 2024 shared task suggest that features and representations derived from item text alone recover only limited information about empirical difficulty \citep{gombert2024predicting,ha-etal-2019-predicting,yaneva2024findings}. The structures that matter for difficulty may therefore extend beyond the most salient or easily named properties of the text. When candidate concepts are generated from item text alone, concept discovery may have difficulty identifying which properties are systematically related to the difficulty experienced by real examinees.

The present study therefore introduces empirical difficulty labels directly into the induction process rather than using them only as a downstream selection criterion. During rubric induction, the model sees both the item content and its empirical difficulty. The resulting principles are intended to capture which item structures are systematically associated with actual examinee performance, rather than simply describing the most salient properties of the text. We then examine whether, and how, explicitly providing these induced difficulty structures changes LLM judgments of new items.

\section{Method}\label{sec3}

\subsection{Datasets}
Two datasets were used: the NCCPA PANCE dataset and the publicly released BEA 2024 shared-task dataset from NBME \citep{yaneva2024findings}. All analyses were conducted independently for the two datasets.

NCCPA PANCE. The dataset contained 528 PANCE items. Item difficulty was represented by the $b$ parameter estimated under a Rasch model, with higher values indicating greater difficulty. Each item included a question stem, response options, and the correct answer. The median stem length was 420 characters. Thirty-five items (6.6\%) contained an \texttt{<img>} tag. The data were split into training and test sets using a 70/30 stratified random split, yielding 370 training items and 158 test items.

BEA 2024 (NBME). The dataset contained 667 retired USMLE clinical items and used the official split of 466 training items and 201 test items. Difficulty was defined as a linear transformation of the proportion of examinees answering the item correctly, with higher values indicating greater difficulty. Each item included a question stem, response options, the correct answer, item type (text only / image-containing), and examination Step (Step 1/2/3). The median stem length was 696 characters. Seventy items (10.5\%) contained images.

\subsection{LLM Rubric Generation}
We used Claude Opus 5 as the common rubric induction model to generate separate rubrics for absolute difficulty judgment and pairwise difficulty comparison. Both rubrics were induced using only training items and their empirical difficulty information. Claude Opus 5 did not see any test items during rubric induction.

For the absolute difficulty rubric, we provided Claude Opus 5 with the full training set in a single call, including 466 USMLE items and 370 PANCE items, together with their observed difficulty. For each item, the model received the item stem, response options, and correct answer, together with the item's empirical difficulty percentile and its membership in one of five bands defined by quintiles of the training-item difficulty distribution (see Table~\ref{tab:bands}). The prompt (Appendix~\ref{app:prompts}) asked the model to characterize each difficulty band in four ways. First, it identified an attribute commonly shared by items in that band but less common in other bands. Second, it identified a subset of items that was not adequately captured by that description. Third, it identified a discernible attribute that varied continuously within the band. Fourth, it selected one representative item from the lower, middle, and upper portions of the band. The entire absolute difficulty rubric was generated in a single model call, with no iterative refinement and no manual selection among multiple generations.

\begin{table}[htbp]
\caption{Difficulty band boundaries and item counts}\label{tab:bands}
\begin{tabular}{@{}cl rrrr rrrr@{}}
\toprule
 & & \multicolumn{4}{c}{USMLE} & \multicolumn{4}{c}{PANCE} \\
\cmidrule(lr){3-6}\cmidrule(lr){7-10}
 & & \multicolumn{2}{c}{Difficulty} & \multicolumn{2}{c}{$n$}
   & \multicolumn{2}{c}{Difficulty} & \multicolumn{2}{c}{$n$} \\
\cmidrule(lr){3-4}\cmidrule(lr){5-6}\cmidrule(lr){7-8}\cmidrule(lr){9-10}
Band & Percentile & Lower & Upper & Train & Test & Lower & Upper & Train & Test \\
\midrule
1 & 0--20   & 0.02 & 0.20 & 90 & 38 & $-2.87$ & $-0.78$ & 74 & 31 \\
2 & 20--40  & 0.20 & 0.34 & 92 & 38 & $-0.78$ & $-0.13$ & 74 & 32 \\
3 & 40--60  & 0.34 & 0.50 & 89 & 33 & $-0.13$ & $0.58$  & 74 & 32 \\
4 & 60--80  & 0.50 & 0.78 & 98 & 51 & $0.58$  & $1.11$  & 74 & 31 \\
5 & 80--100 & 0.78 & 1.38 & 97 & 41 & $1.11$  & $2.75$  & 74 & 32 \\
\midrule
\multicolumn{2}{@{}l}{Total} & & & 466 & 201 & & & 370 & 158 \\
\botrule
\end{tabular}
\end{table}

For rubric induction, we first identified pairs of training items whose empirical difficulty differed by at least 1.5 standard deviations. In the USMLE training set, difficulty ranged from 0.02 to 1.38 ($M = 0.48$, $SD = 0.308$), so this threshold corresponded to a minimum difficulty difference of 0.462. In the PANCE training set, the IRT $b$ parameter ranged from $-2.87$ to 2.75 ($M = 0.16$, $SD = 1.042$), corresponding to a minimum $b$ difference of 1.564. This relatively large minimum gap was intentional. It was used to provide a clear empirical ordering during rubric induction and to reduce label noise and ambiguity from pairs with very similar difficulty. The items represented in the selected pairs covered the full range of item difficulty in both training sets. Their difficulty gaps ranged from 1.56 to 3.38 standard deviations for USMLE and from 1.50 to 3.14 standard deviations for PANCE, with median gaps of 1.75 and 1.71 standard deviations, respectively.

From the eligible pairs, we selected 250 pairs per dataset using a two-step procedure that prioritized item coverage. First, we computed a maximum-cardinality matching to obtain as many pairs as possible without reusing items. We then filled the remaining slots by first selecting pairs that introduced two previously unused items, followed by pairs that introduced one new item, and finally pairs in which both items had already been used. No item was allowed to appear in more than two selected pairs. The resulting 250 pairs covered 411 of the 466 USMLE training items and 366 of the 370 PANCE training items.

Before pairwise comparison rubric induction, Claude Opus 5 judged these training pairs without receiving a rubric or any empirical difficulty information. Each pair was judged independently in both presentation orders, and the comparisons were then classified into three groups. Pairs judged correctly in both orders were classified as stable correct, pairs judged incorrectly in both orders were classified as stable wrong, and pairs for which the judgment changed after reversing the presentation order were classified as inconsistent. The reasoning generated by Claude Opus 5 during these judgments was retained as supporting information for rubric induction.

The pairwise rubric was induced separately for each examination using two model calls. Before the first call, Claude Opus 5 was not given any initial rubric or candidate principles. In the first call, the model received all stable-correct pairs, including the full item content, empirical difficulty percentiles, and the reasoning retained from its earlier judgments of those pairs. It was asked to identify recurring ways of judging difficulty that were supported by the item content and empirical ordering. This call produced a preliminary working account rather than a final rubric.

In the second call, the working account was provided together with all stable-wrong and order-dependent pairs in a single prompt. The model was first asked to examine the stable errors and determine which parts of the working account should be retained, qualified, narrowed, revised, or removed. It then examined the order-dependent pairs to identify reasoning that changed with presentation order and whether those patterns required limits on how the rubric should be applied. New principles could be added only when recurring stable errors revealed a difficulty-related factor that was not covered by the working account. Order dependence itself was not used to generate new difficulty principles and was used only to constrain how existing principles were applied. The final pairwise comparison rubric was then fixed and remained unchanged during subsequent testing. The prompts for both stages are provided in Appendix~\ref{app:prompts}.

\subsection{LLM Judges}
For absolute difficulty judgment, 12 LLMs (7 open-weight and 5 commercial models; Table~\ref{tab:judges}) independently rated each item on a 0--100 difficulty percentile scale under unguided and rubric-guided conditions. Models were instructed to analyze each item and judge its difficulty relative to the target examinee population, without additional information about examinee ability or background.

For pairwise comparison, the same 7 open-weight models judged which of two items would be more difficult for examinees under both conditions. Each pair was evaluated independently in both presentation orders to assess judgment accuracy and sensitivity to item order. The judgment prompts are provided in Appendix~\ref{app:prompts}.

\begin{table}[htbp]
\caption{LLMs used as difficulty judges}\label{tab:judges}
\small
\setlength{\tabcolsep}{2pt}
\begin{tabular}{@{}lllll@{}}
\toprule
Model & Parameters & Architecture & Developer & Task \\
\midrule
\multicolumn{5}{@{}l}{\textit{Open-weight}} \\
Nemotron 3 Super 120B & 120.6B / 12.7B active & Hybrid MoE & NVIDIA    & Absolute, pairwise \\
GPT-OSS 120B          & 117B / 5.1B active    & MoE        & OpenAI    & Absolute, pairwise \\
Llama 3.3 70B         & 70B                   & Dense      & Meta      & Absolute, pairwise \\
Gemma 4 31B           & 30.7B                 & Dense      & Google    & Absolute, pairwise \\
Nemotron 3 Nano 30B   & 31.6B / 3.6B active   & Hybrid MoE & NVIDIA    & Absolute, pairwise \\
MedGemma 27B          & 27B                   & Dense      & Google    & Absolute, pairwise \\
Gemma 3 27B           & 27B                   & Dense      & Google    & Absolute, pairwise \\
\midrule
\multicolumn{5}{@{}l}{\textit{Commercial}} \\
GPT-5.6 Sol           & \multicolumn{2}{c}{Not disclosed} & OpenAI    & Absolute \\
GPT-5.6 Terra         & \multicolumn{2}{c}{Not disclosed} & OpenAI    & Absolute \\
GPT-5.6 Luna          & \multicolumn{2}{c}{Not disclosed} & OpenAI    & Absolute \\
Claude Sonnet 5       & \multicolumn{2}{c}{Not disclosed} & Anthropic & Absolute \\
Claude Opus 5$^{\mathrm{a}}$ & \multicolumn{2}{c}{Not disclosed} & Anthropic & Absolute \\
\botrule
\end{tabular}
\footnotetext{$^{\mathrm{a}}$Claude Opus 5 was also used to induce the difficulty rubrics.}
\end{table}

\subsection{Evaluation}

The evaluation of absolute difficulty judgment was conducted on the held-out test sets, which included 201 USMLE items and 158 PANCE items. We evaluated performance from several perspectives, including use of the 0 to 100 difficulty scale, Spearman rank correlation between model ratings and empirical difficulty, and quadratic weighted kappa (QWK) based on discretized difficulty levels. For QWK, the empirical difficulty of each test item was converted to a percentile relative to the training-item difficulty distribution and then assigned to one of five bands using cut points at the 20th, 40th, 60th, and 80th percentiles. Test items falling outside the observed training range were assigned to the corresponding extreme band. Model-predicted percentiles were discretized using the same 20-point intervals.

The pairwise evaluation used 1,200 item pairs that did not overlap with the 250 pairs used for rubric induction. These pairs were stratified by their true difficulty gap to provide coverage across comparisons with smaller and larger difficulty differences. For each model, the unguided and rubric-guided conditions used exactly the same 1,200 pairs and the same two presentation orders. The primary pairwise metrics were overall accuracy, accuracy among order-consistent judgments and order consistency. Overall accuracy was calculated across all judgments from both presentation orders.
\section{Results}\label{sec4}

\subsection{LLM-Induced Rubrics}
Before examining performance effects, we first assessed whether the judges appeared to use the provided rubric in their reasoning. For absolute difficulty judgments, explicit references to difficulty bands were virtually absent in the unguided condition, with band-related language appearing in only 0\% to 0.7\% of rationales across models. Under rubric guidance, band references increased substantially. Eleven of the 12 judges explicitly referred to a difficulty band in 82.1\% to 100\% of their rationales. The purpose of comparing band-related language was to identify a process-level marker that distinguished the two conditions. Because the band structure was introduced only under rubric guidance while the required output format remained the same, the emergence of band-related language provided evidence that the rubric-induced structure was reflected in the models’ reasoning. We also manually inspected rationales from both conditions to examine how rubric-related criteria were reflected in the models’ reasoning. Illustrative examples are provided in Appendix~\ref{app:examples}.

Across the four induced rubrics (see Appendices~\ref{app:absolute} and~\ref{app:pairwise}), surface complexity was generally not treated as a primary source of medical item difficulty. Both pairwise rubrics explicitly downweighted stem length, number of response options, and the mere presence of images. The PANCE pairwise rubric also indicated that the amount of laboratory data and the apparent number of reasoning steps should not determine difficulty. Similarly, the USMLE absolute rubric found that surface features such as stem length and the presence of laboratory tables did not track difficulty.

Beyond these surface features, PANCE and USMLE shared several basic difficulty structures, including converging cues toward the key, misleading cues, and option competition, although these took somewhat different forms across the two examinations. The rubrics also highlighted some exam-specific patterns. Non-derivable knowledge retrieval was particularly prominent in PANCE, whereas the USMLE rubrics placed greater emphasis on integrating multiple clinical cues.

\subsubsection{PANCE}

\begin{enumerate}
    \item \textbf{Converging cues toward the key:} Items tend to be easier when the cues in the stem clearly establish the disease or clinical context and point directly and consistently to a single, highly typical associated feature. One example in the absolute rubric describes a man with six months of daytime drowsiness and two months of loud snoring. The question asks which finding would most strongly support obstructive sleep apnea. The options are BMI $>$ 30 (correct answer), hepatojugular reflux, nocturnal myoclonus, pectus excavatum, and a systolic ejection murmur. The rubric explains that daytime drowsiness, snoring, and the fact that obstructive sleep apnea is named in the question all point in the same direction. The examinee only needs to recall a highly typical feature associated with the condition, and obesity is one of the best known risk factors for obstructive sleep apnea. The task is therefore quite direct. The stem leaves little uncertainty about the clinical context, and the examinee mainly needs to make a familiar association.
    \item \textbf{Misleading cues and target shift:} Difficulty increases when the most salient cue in the stem is not the information that ultimately determines the answer. This can take the form of a misleading cue in the stem or a shift in the actual target of the question. One example from Band 5 of the absolute rubric describes a binge-drinking woman who is markedly somnolent and responds to questions only by grimacing. Her pupils are reactive, her gag reflex is present, and her cardiopulmonary examination is normal. The question asks for the most appropriate immediate intervention. The options include disulfiram, four-point restraints, endotracheal intubation, fluid resuscitation (correct answer), and nasogastric lavage. The rubric explains that the patient’s obtunded appearance strongly pulls the examinee toward airway protection. To reach the correct answer, the preserved gag reflex has to be given greater weight and must override that initial response.
    
    The pairwise rubric identifies a related pattern in which the target of the question shifts. For example, the patient may have a particular disease, but the question asks about prophylaxis for the patient’s contacts. In another case, the stem may present a visible lesion while the question actually asks about the underlying condition. If the examinee follows only the most obvious clinical content in the stem, it is easy to answer the wrong question. In both misleading-cue and target-shift items, the most prominent information in the stem does not directly correspond to the information on which the final answer depends.

    \item \textbf{Non-derivable knowledge retrieval:} Some items are difficult not because the reasoning itself is complex, but because they require a specific and relatively uncommon piece of knowledge that cannot be derived from the case alone. Band 4 of the absolute rubric describes this pattern clearly. The clinical presentation may be relatively transparent, but the answer depends on specialized knowledge that is not part of the standard presentation of the condition, such as a specific regulatory number, a nerve level, a drug interaction, an uncommon association, or a particular contraindication. For example, one item describes a woman requesting a copper IUD whose history includes endometriosis, prior DVT, prior ectopic pregnancy, prior chlamydia, and Wilson disease. The question asks which condition is a contraindication to a copper IUD, and the correct answer is Wilson disease. The rubric explains that the answer depends on a rarely taught, copper-specific contraindication. In other words, the case itself does not require a complex chain of reasoning. The key issue is whether the examinee knows this specific association involving Wilson disease and copper. Without that knowledge, the answer is difficult to infer from the case alone. At the same time, alternatives such as DVT, ectopic pregnancy, and chlamydia may appear more familiar as reproductive health red flags and therefore provide additional distraction.
    
    The pairwise rubric similarly emphasizes retrieval of specific knowledge. Some items contain no full patient vignette, or only a one-line patient description, and directly ask about a drug mechanism, the pathophysiologic mechanism of a disease or shock state, what a hormone or physiologic system responds to, a definition, a diagnostic criterion, a category label, an ethical or legal principle, or a fixed numerical or eligibility threshold. The rubric specifically cautions against treating these items as easier simply because the fact seems elementary to the LLM. Similarly, some items mainly require matching a description to the correct term or choosing among adjacent terms within the same vocabulary family, such as surgical complications, dementia subtypes, shock categories, anxiety disorders, renal syndromes, or ethical principles. Even when the stem gives a description close to a textbook definition, the item is not necessarily easy.

    \item \textbf{Option competition:} Difficulty also arises when multiple options remain defensible. Items become harder when several options are clinically plausible, or when multiple statements are actually true but the examinee must determine which is most important, greatest in magnitude, most appropriate for the patient, or highest in priority. One example in the absolute rubric describes a patient who has been taking warfarin and whose INR had previously been stable. The INR suddenly increases to 9.0, the patient has no bleeding, and the patient reports accidentally taking a double dose two days earlier. The options include advising the patient to eat more green leafy vegetables, withholding warfarin that evening (correct answer), checking partial thromboplastin time, lowering the daily warfarin dose, and referring the patient to a hematologist. The rubric explains that the keyed answer is a minimal, self-limiting action that competes with several plausible active interventions. The difficulty comes from the fact that several management options appear reasonable. The examinee must consider the current severity, the patient’s previously stable treatment, and the clearly temporary cause of the abnormal INR in order to determine which action is most appropriate in this specific situation.
    
    The pairwise rubric identifies a similar pattern in which two or more options may genuinely apply to the patient, while the keyed answer wins only by degree. For example, two drug classes may both raise a particular level, two adverse effects may both occur, or two options may both be real risk factors, while the question asks which has the greater effect, which is the most important, or which is the most likely. In such cases, simply determining whether an option is correct is not enough. The examinee must further compare the degree, importance, or priority of multiple true or defensible alternatives.
\end{enumerate}

\subsubsection{USMLE}

\begin{enumerate}
    \item \textbf{Converging cues toward the key:} This pattern was similar to PANCE. The easiest USMLE items also tended to contain hallmark findings that pointed directly to the keyed answer, while the other options came from clearly different disease categories or organ systems. Once the key finding was recognized, the correct answer could be identified with little additional work. For example, one item described a patient with chest pain that worsened with inspiration and improved when leaning forward, along with fever, distant heart sounds, and diffuse ST elevation. The question asked which tissue was injured, and the correct answer was the pericardium. Multiple findings in the stem converged on the same clinical problem, while alternatives such as the aortic intima, esophageal sphincter, myocardium, and pleura could not explain the full presentation.

    \item \textbf{Misleading cues:} This was a particularly prominent source of difficulty in USMLE. Compared with the broader salience to relevance mismatch seen in PANCE, misleading cues in USMLE often took a more specific form. The stem explicitly provided a fact that supported a particular wrong answer. For example, one patient with heart failure was taking lisinopril, spironolactone, digoxin, and metoprolol. The patient developed weakness, with a potassium level of 7.3, urea of 40, creatinine of 1.8, and a normal urinalysis. The correct answer was an adverse effect of medications, but the abnormal renal function values also provided genuine support for renal failure as an alternative. The difficulty therefore did not come simply from a distractor that looked plausible. Specific information in the stem actually supported another answer. At still higher levels of difficulty, the rubric described cases in which a specific value or statement in the stem appeared to argue directly against the keyed answer and had to be overridden.

    \item \textbf{Option competition:} This source of difficulty also appeared in PANCE, but it took a somewhat different form in USMLE. In PANCE, competition more often arose because several answers were clinically defensible in the current case. In USMLE, the options were more often members of the same knowledge family, so recognizing the disease still did little to eliminate them. The pairwise rubric gave examples such as two muscles from the same muscle group, two disorders affecting the same organ, different forms of the same drug class, different parts of the same structure, or two nearly synonymous descriptions of a mechanism. The challenge was therefore to distinguish among near neighbors within the same knowledge domain.

    \item \textbf{Inferential demand:} This source of difficulty was more explicit and systematic in the USMLE rubrics than in PANCE. The absolute rubric showed a gradual progression. In easier items, stem cues could map directly onto the correct answer. At the next level, the disease itself might still be easy to recognize, but the question required one additional step, such as identifying a mediator, microbial property, metabolite, vessel, complication, or treatment. At higher levels, no single piece of information was sufficient. Multiple elements of the stem had to be integrated before options could be eliminated, such as a medication together with a laboratory trend, an exposure together with a time course, or a value together with a response to an intervention. As difficulty increased, the clinical information in the stem became less directly connected to the final answer and required more information integration and inferential progression.
\end{enumerate}

\subsection{Effect of Rubric Guidance on Absolute Difficulty Judgments}
Tables~\ref{tab:scale-usmle} and~\ref{tab:scale-pance} show that rubric guidance clearly changed how the models used the difficulty percentile scale. All models used more distinct percentile values in the guided condition, suggesting that the rubric not only expanded scale use for some models but also reduced repeated use of a small set of preferred scores, allowing more differentiated use of the 0 to 100 scale. At the same time, the standard deviation of predicted difficulty increased under rubric guidance for most models, further indicating broader use of the difficulty scale.

The unguided condition also revealed different forms of scale compression across models. The GPT-5.6 models showed especially strong compression at the high end of the scale. On USMLE, the maximum predicted percentile was 68 for Sol, Terra, and Luna. On PANCE, the corresponding maxima were only 58, 72, and 64. This suggests that without the rubric, these models rarely classified items as extremely difficult. With rubric guidance, the maximum predictions for these models generally increased to between 86 and 92 across the two datasets. Gemma 4 showed compression toward the middle of the scale, with limited use of both the low and high ends. In the unguided condition, it used only 10 distinct percentile values on USMLE and 8 on PANCE, with predictions restricted to a range of 10 to 65 on both datasets. Under rubric guidance, the number of distinct values increased to 35 on both datasets. Although this remained relatively low, the prediction range expanded substantially. Gemma 3 showed a different pattern. In the unguided condition, its minimum prediction was 20 on both USMLE and PANCE, indicating compression at the low end, while its maximum predictions were relatively high. With rubric guidance, the prediction range expanded in both directions. However, changes in the mean varied across models, suggesting that the primary effect of rubric guidance was to broaden and differentiate the prediction distribution rather than systematically shift overall difficulty judgments in a single direction.

\begin{table}[htbp]
\caption{Use of the difficulty percentile scale on USMLE ($n = 201$)}\label{tab:scale-usmle}
\small
\setlength{\tabcolsep}{4pt}
\begin{tabular}{@{}lrrrrrrrr@{}}
\toprule
 & \multicolumn{4}{c}{Unguided} & \multicolumn{4}{c}{Rubric-guided} \\
\cmidrule(lr){2-5}\cmidrule(lr){6-9}
Judge & $M$ & $SD$ & Distinct & Range & $M$ & $SD$ & Distinct & Range \\
\midrule
Claude Opus 5             & 25.2 & 18.3 & 45 & 2--76  & 38.8 & 23.8 & 73 & 4--86 \\
GPT-5.6 Sol               & 19.3 & 12.1 & 33 & 2--68  & 30.6 & 20.7 & 52 & 2--88 \\
GPT-5.6 Terra             & 24.5 & 11.0 & 21 & 4--68  & 29.5 & 18.3 & 48 & 3--86 \\
Claude Sonnet 5           & 39.1 & 14.2 & 16 & 15--72 & 43.4 & 26.7 & 46 & 3--91 \\
Gemma 4 31B               & 29.3 & 11.6 & 10 & 10--65 & 30.1 & 21.0 & 35 & 0--92 \\
GPT-5.6 Luna              & 17.5 &  9.0 & 23 & 2--68  & 26.2 & 21.4 & 54 & 2--91 \\
Nemotron 3 Super 120B     & 25.5 & 13.9 & 26 & 0--73  & 28.4 & 23.1 & 37 & 0--99 \\
GPT-OSS 120B              & 29.6 & 17.2 & 23 & 5--78  & 22.4 & 17.5 & 44 & 2--96 \\
Gemma 3 27B               & 53.2 & 16.1 & 13 & 20--85 & 54.1 & 21.8 & 32 & 8--92 \\
Llama 3.3 70B             & 54.6 & 21.5 & 11 & 20--85 & 38.3 & 22.5 & 33 & 5--95 \\
Nemotron 3 Nano 30B       & 32.6 & 18.4 & 27 & 0--85  & 20.3 & 23.0 & 43 & 0--99 \\
MedGemma 27B              & 49.4 & 13.9 & 17 & 0--75  & 37.7 & 21.0 & 44 & 0--99 \\
\botrule
\end{tabular}
\footnotetext{\textit{Note.} Distinct = number of unique percentile values used.}
\end{table}

\begin{table}[htbp]
\caption{Use of the difficulty percentile scale on PANCE ($n = 158$)}\label{tab:scale-pance}
\small
\setlength{\tabcolsep}{4pt}
\begin{tabular}{@{}lrrrrrrrr@{}}
\toprule
 & \multicolumn{4}{c}{Unguided} & \multicolumn{4}{c}{Rubric-guided} \\
\cmidrule(lr){2-5}\cmidrule(lr){6-9}
Judge & $M$ & $SD$ & Distinct & Range & $M$ & $SD$ & Distinct & Range \\
\midrule
Claude Opus 5            & 20.1 & 18.3 & 46 & 2--82  & 25.8 & 22.3 & 57 & 2--90 \\
GPT-5.6 Sol               & 12.2 &  9.9 & 27 & 1--58  & 20.0 & 19.4 & 44 & 0--92 \\
GPT-5.6 Terra             & 18.3 &  9.6 & 22 & 4--72  & 24.8 & 19.5 & 48 & 1--87 \\
Claude Sonnet 5           & 29.6 & 14.0 & 17 & 5--68  & 26.1 & 23.3 & 40 & 2--86 \\
Gemma 4 31B               & 22.2 &  8.2 &  8 & 10--65 & 26.0 & 16.8 & 35 & 2--88 \\
GPT-5.6 Luna              & 12.0 &  7.9 & 18 & 2--64  & 26.4 & 21.9 & 44 & 0--87 \\
Nemotron 3 Super 120B     & 18.1 & 11.8 & 24 & 4--70  & 17.8 & 15.7 & 36 & 0--86 \\
GPT-OSS 120B              & 20.8 & 13.5 & 19 & 5--65  & 18.3 & 17.9 & 36 & 2--92 \\
Gemma 3 27B               & 46.6 & 18.6 & 13 & 20--80 & 43.4 & 22.3 & 28 & 5--88 \\
Llama 3.3 70B             & 39.5 & 21.0 &  9 & 20--85 & 30.4 & 20.0 & 28 & 10--92 \\
Nemotron 3 Nano 30B       & 21.8 & 14.0 & 23 & 0--65  & 10.7 & 13.1 & 34 & 0--61 \\
MedGemma 27B              & 44.2 & 13.9 & 18 & 0--75  & 24.0 & 19.7 & 36 & 0--100 \\
\botrule
\end{tabular}
\footnotetext{\textit{Note.} Distinct = number of unique percentile values used.}
\end{table}

Table~\ref{tab:qwk} presents coarse-grained agreement across the five predefined difficulty bands. Quadratic weighted kappa (QWK) was used to assess agreement between the predicted and observed bands while accounting for the ordered nature of the scale. Overall, rubric guidance produced mostly positive changes, although the effects varied across models. Claude Opus 5 showed the most notable change. As the model used to induce the difficulty rubric, Opus 5 was not the strongest judge in the unguided condition, with QWK values below those of Claude Sonnet 5 on both USMLE and PANCE. With rubric guidance, however, Opus 5 achieved the highest QWK on both datasets, reaching .326 and .245, respectively. The GPT-5.6 models also showed a relatively consistent pattern of improvement, with Sol, Terra, and Luna all showing higher QWK under rubric guidance on both datasets. In contrast, the open-weight models showed more heterogeneous results. One exception to this cross-dataset inconsistency was Gemma 4, which showed higher QWK under rubric guidance on both datasets. Other models often showed dataset-specific changes. For example, Llama 3.3 70B increased from .123 to .226 on USMLE but decreased from .228 to .121 on PANCE. Overall, the effect of rubric guidance on coarse-grained difficulty judgments varied substantially across both models and datasets.

\begin{table}[htbp]
\caption{Quadratic weighted kappa (QWK) for agreement between predicted and observed five-level difficulty bands on the held-out test sets}\label{tab:qwk}
\small
\begin{tabular}{@{}lcccccc@{}}
\toprule
& \multicolumn{3}{c}{USMLE ($n=201$)} & \multicolumn{3}{c}{PANCE ($n=158$)} \\
\cmidrule(lr){2-4}\cmidrule(lr){5-7}
Judge & Unguided & Guided & $\Delta$ & Unguided & Guided & $\Delta$ \\
\midrule
Claude Opus 5         & .196 & .326 & $+$.130 & .162 & .245 & $+$.082 \\
GPT-5.6 Sol            & .107 & .173 & $+$.066 & .055 & .172 & $+$.117 \\
GPT-5.6 Terra          & .148 & .164 & $+$.015 & .125 & .178 & $+$.053 \\
Claude Sonnet 5        & .239 & .252 & $+$.013 & .229 & .219 & $-$.010 \\
Gemma 4 31B            & .142 & .236 & $+$.094 & .099 & .147 & $+$.048 \\
GPT-5.6 Luna           & .076 & .087 & $+$.012 & .031 & .174 & $+$.142 \\
Nemotron 3 Super 120B  & .128 & .087 & $-$.041 & .099 & .063 & $-$.036 \\
GPT-OSS 120B           & .083 & .135 & $+$.052 & .135 & .038 & $-$.096 \\
Gemma 3 27B            & .099 & .041 & $-$.058 & .141 & .150 & $+$.009 \\
Llama 3.3 70B          & .123 & .226 & $+$.103 & .228 & .121 & $-$.107 \\
Nemotron 3 Nano 30B    & .072 & $-$.011 & $-$.082 & .028 & .046 & $+$.018 \\
MedGemma 27B           & .055 & .093 & $+$.038 & .131 & .117 & $-$.014 \\
\botrule
\end{tabular}
\footnotetext{\textit{Note.} Judges are ordered by unguided Spearman correlation on USMLE (Table~\ref{tab:pointwise}). $\Delta$ = Guided $-$ Unguided, computed from unrounded values.}
\end{table}

\begin{table}[htbp]
\caption{Spearman correlations between LLM-estimated and observed item difficulty on the held-out test sets}\label{tab:pointwise}
\small
\begin{tabular}{@{}lcccccc@{}}
\toprule
& \multicolumn{3}{c}{USMLE ($n=201$)} & \multicolumn{3}{c}{PANCE ($n=158$)} \\
\cmidrule(lr){2-4}\cmidrule(lr){5-7}
Judge & Unguided & Guided & $\Delta$ & Unguided & Guided & $\Delta$ \\
\midrule
Claude Opus 5          & .443 & .436 & $-$.006 & .488 & .476 & $-$.011 \\
GPT-5.6 Sol            & .394 & .352 & $-$.042 & .421 & .391 & $-$.030 \\
GPT-5.6 Terra          & .389 & .286 & $-$.103 & .427 & .361 & $-$.066 \\
Claude Sonnet 5        & .385 & .291 & $-$.094 & .453 & .387 & $-$.067 \\
Gemma 4 31B            & .315 & .359 & $+$.044 & .403 & .364 & $-$.039 \\
GPT-5.6 Luna           & .297 & .212 & $-$.085 & .405 & .350 & $-$.055 \\
Nemotron 3 Super 120B  & .222 & .109 & $-$.113 & .308 & .234 & $-$.074 \\
GPT-OSS 120B           & .193 & .301 & $+$.108 & .339 & .168 & $-$.171 \\
Gemma 3 27B            & .160 & .066 & $-$.093 & .186 & .235 & $+$.049 \\
Llama 3.3 70B          & .134 & .267 & $+$.133 & .317 & .272 & $-$.045 \\
Nemotron 3 Nano 30B    & .129 & $-$.022 & $-$.151 & .023 & $-$.095 & $-$.118 \\
MedGemma 27B           & .102 & .154 & $+$.052 & .212 & .216 & $+$.004 \\
\botrule
\end{tabular}
\footnotetext{\textit{Note.} Judges are ordered by unguided Spearman correlation on USMLE. $\Delta$ = Guided $-$ Unguided, computed from unrounded values.}
\end{table}

Table~\ref{tab:pointwise} further examines the models’ ability to rank items by difficulty at a fine-grained level. Notably, several commercial models already showed relatively strong difficulty ranking in the unguided condition. On USMLE, the Spearman correlations for Claude Opus 5, GPT-5.6 Sol, GPT-5.6 Terra, and Claude Sonnet 5 were all close to or above .39. These values were higher than the approximately .25 correlation achieved by the winning customized embedding approach in the 2024 BEA shared task organized by NBME \citep{gombert2024predicting}, as well as the approximately .29 correlation reported for the best single feature from the second-place simulated-student approach \citep{duenas2024upn}. This suggests that current frontier LLMs can recover a meaningful amount of item difficulty information through direct judgment alone, without task-specific training.

Performance among the open-weight models varied substantially and did not follow a simple pattern by parameter size. On PANCE, Gemma 4 31B achieved an unguided Spearman correlation of .403, higher than Nemotron 3 Super 120B and GPT-OSS 120B at .308 and .339. Llama 3.3 70B also reached .317, while performance among the remaining open-weight models varied considerably.

Rubric guidance did not lead to better fine-grained ranking overall. All five commercial models showed lower Spearman correlations under rubric guidance on both datasets. Claude Opus 5 changed the least, decreasing from .443 to .436 on USMLE and from .488 to .476 on PANCE. This suggests that it largely preserved its original ranking ability but did not gain further improvement from the rubric it had induced. The declines were larger for the other commercial models. Results for the open-weight models were more heterogeneous. On USMLE, Gemma 4, GPT-OSS 120B, Llama 3.3 70B, and MedGemma improved, with the largest gains observed for Llama 3.3 70B and GPT-OSS 120B. On PANCE, only Gemma 3 and MedGemma improved, and the change for MedGemma was very small. Overall, rubric guidance did not produce a consistent improvement in fine-grained item difficulty ranking, with effects varying across both models and datasets.

One possible explanation is that rubric guidance primarily changed how models mapped items onto the difficulty scale rather than improving their fine-grained ordering. Under the unguided condition, substantial scale compression caused many items to cluster within a relatively narrow portion of the scale. By expanding the distribution of predicted difficulty, the rubric may have moved more items into the appropriate broad difficulty bands, thereby improving QWK. Spearman correlation, however, depends mainly on the relative ordering of items rather than on the spread of predicted scores. Thus, broader scale use could improve coarse-grained agreement without improving, and potentially slightly disrupting, fine-grained ranking.

\subsection{Effect of Rubric Guidance on Pairwise Comparison}
\begin{table}[htbp]
\caption{Pairwise difficulty judgment with and without rubric guidance}\label{tab:pairwise_guidance}
\small
\setlength{\tabcolsep}{4pt}
\begin{tabular}{@{}lccccccccc@{}}
\toprule
 & \multicolumn{3}{c}{Acc$_{\text{all}}$} & \multicolumn{3}{c}{Acc$_{\text{cons}}$} & \multicolumn{3}{c}{Consistency (\%)} \\
\cmidrule(lr){2-4}\cmidrule(lr){5-7}\cmidrule(lr){8-10}
Judge & U & G & $\Delta$ & U & G & $\Delta$ & U & G & $\Delta$ \\
\midrule
\multicolumn{10}{@{}l}{\textit{USMLE}} \\
Gemma 4 31B           & .635 & .645 & $+$.009 & .664 & .677 & $+$.013 & 82.6 & 81.9 & $-$0.7 \\
Nemotron 3 Super 120B & .604 & .550 & $-$.054 & .634 & .571 & $-$.063 & 77.2 & 70.2 & $-$7.0 \\
GPT-OSS 120B          & .584 & .589 & $+$.005 & .604 & .620 & $+$.016 & 81.2 & 74.5 & $-$6.7 \\
MedGemma 27B          & .568 & .590 & $+$.023 & .588 & .628 & $+$.040 & 77.2 & 70.8 & $-$6.4 \\
Gemma 3 27B           & .567 & .573 & $+$.006 & .580 & .594 & $+$.014 & 83.7 & 77.8 & $-$5.9 \\
Llama 3.3 70B         & .565 & .558 & $-$.007 & .586 & .574 & $-$.012 & 75.1 & 78.4 & $+$3.3 \\
Nemotron 3 Nano 30B   & .562 & .584 & $+$.023 & .592 & .624 & $+$.032 & 67.0 & 68.0 & $+$1.0 \\
\midrule
\multicolumn{10}{@{}l}{\textit{PANCE}} \\
Gemma 4 31B           & .632 & .583 & $-$.049 & .656 & .602 & $-$.054 & 84.7 & 80.5 & $-$4.2 \\
GPT-OSS 120B          & .604 & .609 & $+$.005 & .627 & .644 & $+$.017 & 82.3 & 75.7 & $-$6.6 \\
Nemotron 3 Super 120B & .598 & .575 & $-$.023 & .630 & .605 & $-$.025 & 75.8 & 71.2 & $-$4.6 \\
Nemotron 3 Nano 30B   & .572 & .563 & $-$.009 & .615 & .595 & $-$.020 & 62.7 & 66.5 & $+$3.8 \\
MedGemma 27B          & .554 & .534 & $-$.020 & .569 & .547 & $-$.022 & 78.2 & 72.1 & $-$6.1 \\
Llama 3.3 70B         & .527 & .534 & $+$.008 & .535 & .549 & $+$.014 & 75.3 & 69.2 & $-$6.1 \\
Gemma 3 27B           & .522 & .538 & $+$.016 & .527 & .555 & $+$.028 & 81.9 & 69.2 & $-$12.7 \\
\botrule
\end{tabular}
\footnotetext{\textit{Note.} U = unguided; G = rubric-guided; $\Delta$ = G $-$ U, computed from unrounded values. Acc$_{\text{all}}$ = accuracy over all pairs, with order-inconsistent judgments scored as 0.5; Acc$_{\text{cons}}$ = accuracy over order-consistent pairs; Consistency = percentage of pairs judged the same in both presentation orders.}
\end{table}

As shown in Table~\ref{tab:pairwise_guidance}, unguided pairwise difficulty judgment did not improve monotonically with model size. Gemma 4 31B achieved the highest overall accuracy on both USMLE and PANCE, at .635 and .632, respectively. It also showed high order consistency, ranking first on PANCE and second on USMLE.
The effects of rubric guidance varied substantially across models and examinations. On USMLE, all four smaller models in the 27B to 31B range improved in overall accuracy, with gains ranging from .006 to .023, whereas the effects among the 70B to 120B models were mixed. This pattern did not replicate on PANCE, where three of the four models in the 27B to 31B range showed lower accuracy under rubric guidance. This suggests that an explicit rubric may provide useful structure for smaller models under some conditions, but its benefits are not consistent and appear to depend on how well the induced guidance transfers to new item comparisons.

Compared with accuracy, order consistency showed a more consistent overall decline under rubric guidance. One possible explanation is that the rubric introduced multiple explicit considerations into the comparison process. When different criteria favored different items, their relative salience may have shifted with presentation order, making the resulting judgments less stable.

\section{Discussion}\label{sec5}
One major finding of this study is that medical item difficulty appears to be better understood as a relational property than as a property of any isolated item component. Stem length, number of response options, and the presence of images did not consistently explain difficulty. Instead, the recurring patterns concerned how information in the stem related to the keyed answer and competing options. A cue could reduce difficulty by narrowing the answer space, but the same type of cue could increase difficulty if it supported a plausible competing option. Similarly, recognizing the underlying condition did not necessarily make an item easy. What also mattered was what remained to be done between recognizing the clinical problem and arriving at the final answer. From this perspective, medical item difficulty depends less on what an item contains than on how its information structures the examinee’s path to an answer. The two examinations were broadly similar in this respect, although some differences emerged in how these structures were expressed.

The effects of rubric guidance further suggest that making difficulty structure explicit and improving difficulty judgment are not the same problem. In absolute judgment, the clearest effect of the rubric was to change how models organized and used the difficulty scale, whereas its effects on fine-grained relative ranking were inconsistent. In pairwise comparison, all models in the 27B to 31B range improved on USMLE, but this pattern did not replicate on PANCE. Thus, across both absolute and pairwise judgment, explicitly providing the rubric did not produce consistent performance gains.
One possible explanation is that some of the rules captured by the rubrics were already reflected in the models’ unguided judgments. This possibility was particularly apparent in absolute judgment. Several commercial models already showed relatively strong difficulty ranking without rubric guidance, yet explicit guidance did not further improve their fine-grained ranking. Even Claude Opus 5, the model that induced the rubrics, did not improve when using its own rubric. This suggests that a natural-language rubric may represent only an incomplete externalization of a model’s original judgment process. It can compress some implicit structure into principles that are readable and inspectable, but those explicit principles may not capture all of the fine-grained information used in the model’s original judgments. The value of rubric induction therefore should not be judged solely by whether it further improves predictive performance.

At the same time, a more interpretable judgment framework is not necessarily a more accurate or stable one. The decline in order consistency in pairwise comparison suggests that an explicit rubric, while providing additional structure, also requires the model to weigh multiple rules. Presentation order may therefore change the salience of different pieces of evidence during reasoning. In other words, adding explicit judgment principles does not necessarily reduce uncertainty and may sometimes increase the amount of information that the model must coordinate.

Overall, the value of label-guided rubric induction may lie less in providing a reliable method for improving LLM difficulty judgments than in transforming statistical differences in empirical item difficulty into structural hypotheses that can be read, compared, and tested. It provides a path from empirical labels to interpretable accounts of difficulty. Future work could examine these principles through expert review, test whether different models independently recover similar structures, and evaluate how well USMLE and PANCE rubrics transfer across examinations.

This study has several limitations. All four rubrics were induced by Claude Opus 5, so the resulting structures may partly reflect model-specific analytical tendencies, and they have not yet been independently evaluated by medical education or item-writing experts. The study also included only USMLE and PANCE, so it remains unclear whether the identified structures generalize to other medical examinations. In addition, the pairs used for pairwise rubric induction all had relatively large empirical difficulty differences. Although this design reduced ambiguity in the supervision signal, it may also have produced principles that are better suited to distinguishing clearly different items than to resolving subtle differences between items of similar difficulty. Finally, the rubrics describe recurring patterns associated with empirical difficulty rather than causal mechanisms established through experimental manipulation.

\backmatter

\bmhead{Acknowledgements}

We thank the National Commission on Certification of Physician
Assistants (NCCPA) for providing the PANCE data used in this
study, and the National Board of Medical Examiners (NBME)
and the organizers of the BEA 2024 shared task for making
the USMLE item data publicly available.

\bibliography{sn-bibliography}

\begin{appendices}

\section{Prompt Templates}\label{app:prompts}

This appendix provides the prompt templates used for rubric induction and
difficulty judgment. Bracketed text indicates dataset-specific content or
content inserted dynamically during model calls. For USMLE,
\texttt{[EXAMINATION]} refers to the United States Medical Licensing
Examination (USMLE), and the target population is students at US medical
schools. For PANCE, \texttt{[EXAMINATION]} refers to the Physician Assistant
National Certifying Examination (PANCE), and the target population is
graduating students of US physician assistant programs.

\subsection{Absolute Difficulty Rubric Induction}
\label{sec:absolute-rubric-prompt}

For each dataset, Claude Opus 5 received the labeled training items and the
following system prompt. Each item was supplied with its empirical difficulty
percentile and its membership in one of five equal-sized difficulty bands.

\begin{Verbatim}[fontsize=\small,breaklines=true,breaksymbolleft={},frame=single,framesep=6pt,rulecolor=\color{gray}]
You are an experienced medical examination item writer and psychometrician. The items below are from the [EXAMINATION], taken by [TARGET POPULATION].

Each item is shown with its empirical difficulty percentile among the training items and with the band it falls in. The 5 bands are equal-sized quintiles of the training difficulty distribution, band 1 being the easiest.

Characterize each band in four ways.

First, an attribute commonly shared by items in that band and less common in the other bands. Go by what the items and the difficulty data show, not by what usually makes examination items hard.

Second, a subset of items in the band that your description does not adequately capture. Name it and say what those items are like instead.

Third, an attribute that varies continuously within the band, so that a reader can tell an item near the bottom of the band from one near the top.

Fourth, one representative item from the lower, middle and upper portions of the band, each with a short statement of why it sits there.

Say plainly and concretely what distinguishes the band. Do not invent abstract names for the sake of generalising, and do not decide in advance how much to write about each band.

Put the whole of your answer inside a single pair of tags, with nothing before the opening tag and nothing after the closing one.

<rubric>
...
</rubric>
\end{Verbatim}

\subsection{Pairwise Difficulty Rubric Induction}
\label{sec:pairwise-rubric-prompt}

Pairwise rubric induction was conducted in two stages. The first stage used
pairs that Claude Opus 5 had judged correctly under both presentation orders.
The second stage used stable errors and order-dependent pairs to test and
revise the initial account.

\subsubsection*{Stage 1: Initial Induction}

\begin{Verbatim}[fontsize=\small,breaklines=true,breaksymbolleft={},frame=single,framesep=6pt,rulecolor=\color{gray}]
You are working out, from examination items, real examinee difficulty data, and a record of how a language model judged them, what makes one item harder than another on [EXAMINATION].

Every pair was judged twice, once with each item presented first. Only pairs judged correctly both times, choosing the same item, are included here.

Work out which ways of judging repeatedly worked in these correct and stable judgements. You may report a way of judging that the model used repeatedly and that the items and the real difficulty data bear out, and you may also report a regularity you find yourself after comparing many pairs against the real difficulty, which the model's reasoning never states.

Go by the item content and the real difficulty data. The model's reasoning is only what it was thinking at the time; it can be wrong even when the final choice was right.

Do not decide in advance how many there should be. Say plainly and concretely how the judgement was actually made and why it came out right, without inventing abstract names for the sake of generalising.

Put the whole of your answer inside a single pair of tags, exactly as below, with nothing before the opening tag and nothing after the closing one. Give the ways of judging that recur and worked, together with the cases that support each. This is a first account for the next stage to test further, not final guidance for anyone.

<working_account>
...
</working_account>
\end{Verbatim}

\subsubsection*{Stage 2: Revision and Final Guidance}

\begin{Verbatim}[fontsize=\small,breaklines=true,breaksymbolleft={},frame=single,framesep=6pt,rulecolor=\color{gray}]
From pairs that were judged correctly and stably, you have written a first account of which ways of judging repeatedly worked. You are now shown two other kinds of pair.

STABLE ERRORS: the model chose the same item as harder under both presentation orders, but the real examinee data shows the other item was harder.

ORDER-DEPENDENT PAIRS: the model chose a different item as harder under the two presentation orders. There are [N_ORDER_DEPENDENT] of these.

Use these cases to test and revise the account.

Take the stable errors first. Work out which parts of the account still hold, and which need a condition, a narrower scope, a correction, or removal. If the errors repeatedly show something the account does not cover at all, you may add it.

Then take the order-dependent pairs. Compare the reasoning given under the two orders for the same pair and work out which kinds of reason change with presentation order, and whether that means anything in the account needs a limit on how it is used. Do not turn order-dependence itself into a new rule about difficulty.

Go by the item content and the real difficulty data; the model's reasoning helps you see how it was judging. Do not make a large change on the strength of one or two cases.

Say plainly and concretely how to judge, without inventing abstract names for the sake of generalising, and without deciding in advance how many there should be.

Each of the four sections below must appear exactly once, inside its own tags, in this order, with nothing outside them.

<failure_analysis>
What the stable errors and the order-dependent pairs repeatedly show, and the cases that support it.
</failure_analysis>

<revision_notes>
Which parts of the account were kept, changed, limited, removed or added, and why.
</revision_notes>

<synthesis>
What these development cases show overall about judging difficulty on this examination: what works, what goes wrong, and what cannot be judged reliably from the item alone.
</synthesis>

<guidance>
Final guidance for a judge about to compare two new items. Keep only what these cases show is still useful. Tell it plainly how to judge, or what it should stop treating as decisive. Merge anything that repeats; do not decide the number in advance. The downstream task requires a forced choice between the two items; do not introduce a tie or abstention option.

Before finalizing, read the guidance through as a judge would use it and reconcile any parts that cannot all be followed together.

The judge may meet items that also appear among these development cases. Do not include details specific enough to identify an item, such as a distinctive cue-and-answer combination, an exact value, or a named fact tied to a particular case. General medical examples or category-level descriptions are fine.
</guidance>
\end{Verbatim}

\subsection{Absolute Difficulty Judgment}
\label{sec:absolute-judgment-prompt}

The same base prompt was used in the unguided and rubric-guided conditions.
In the rubric-guided condition, the induced absolute difficulty rubric was
inserted at the location marked \texttt{[INDUCED RUBRIC]}. In the unguided
condition, the rubric preamble and induced rubric were omitted.

\begin{Verbatim}[fontsize=\small,breaklines=true,breaksymbolleft={},frame=single,framesep=6pt,rulecolor=\color{gray}]
You are an experienced [EXAMINATION] item writer and psychometrician. Examinees are [TARGET POPULATION].

Difficulty is expressed as a percentile within a reference set of items from this examination. Percentile 0 marks the item that the largest share of examinees answered correctly; percentile 100 marks the item that the smallest share answered correctly. Items are spread evenly across the scale.

Some items refer to figures, photographs, graphs, or table column headers that examinees can see but that are not reproduced in the text you receive. Examinees have access to that material; its absence from your view is not itself a source of difficulty for them.

[GUIDED CONDITION ONLY]
Below is a description of the difficulty structure of this item bank, derived from the observed difficulty of a reference set of items. Use it to place the item: decide which band it belongs to, then place it within that band.

[INDUCED RUBRIC]
[END GUIDED CONDITION]

Estimate the difficulty percentile of the item shown.

Output EXACTLY in this format:
<reasoning> your analysis </reasoning>
<percentile> a whole number from 0 to 100 </percentile>
\end{Verbatim}

The user message supplied to the model was:

\begin{Verbatim}[fontsize=\small,breaklines=true,breaksymbolleft={},frame=single,framesep=6pt,rulecolor=\color{gray}]
[ITEM TEXT]

Estimate the difficulty percentile. Follow the required output format.
\end{Verbatim}

\subsection{Pairwise Difficulty Judgment}
\label{sec:pairwise-judgment-prompt}

The same base prompt was used in the unguided and rubric-guided conditions.
In the rubric-guided condition, the final pairwise guidance produced during
rubric induction was inserted at the location marked
\texttt{[INDUCED RUBRIC]}. In the unguided condition, the rubric preamble and
induced rubric were omitted.

\begin{Verbatim}[fontsize=\small,breaklines=true,breaksymbolleft={},frame=single,framesep=6pt,rulecolor=\color{gray}]
You are a [TARGET EXAMINEE] taking the [EXAMINATION].

You will see two items, A and B, each with its stem, options, and correct answer.

Some items refer to figures, photographs, graphs, or table column headers that examinees can see but that are not reproduced in the text you receive. Examinees have access to that material; its absence from your view is not itself a source of difficulty for them.

[GUIDED CONDITION ONLY]
Below is guidance on judging which of two items is harder, derived from items whose observed difficulty ordering is known. Use it when comparing the two items.

[INDUCED RUBRIC]
[END GUIDED CONDITION]

Judge which item is HARDER for examinees -- meaning fewer examinees answer it correctly.

Compare the two items directly: explain what specifically makes the harder one harder THAN THE OTHER, and where examinees are most likely to fail.

Output EXACTLY in this format:
<reasoning> your direct comparison of the two items </reasoning>
<harder> A or B </harder>
\end{Verbatim}

The user message supplied to the model was:

\begin{Verbatim}[fontsize=\small,breaklines=true,breaksymbolleft={},frame=single,framesep=6pt,rulecolor=\color{gray}]
Question A
[ITEM A]

Question B
[ITEM B]

Which is harder? Follow the required output format.
\end{Verbatim}

\section{Examples of Model Reasoning}\label{app:examples}

\noindent\textit{Note.} The examples below are illustrative excerpts from model reasoning under the unguided and rubric-guided conditions. They are included to show how the reasoning process changed when rubric guidance was provided and are not intended as a systematic qualitative analysis of all model responses.

\subsection{Absolute Difficulty Judgment}

\noindent USMLE item 1, judged by Claude Opus 5.
\begin{description}\small
\item[Unguided (percentile = 17).] Classic urge incontinence vignette; not stress-related, no infection, normal glucose. Straightforward for most examinees. Slight distraction from glucose 120 and bacteria, but answer is clear.
\item[Rubric-guided (percentile = 22).] The stem gives urge-type incontinence not provoked by cough/sneeze, with normal labs excluding infection, glucose, and renal issues. The key (detrusor instability) is essentially the name of urge incontinence, and distractors are actively negated by the stem's data (glucose 120 borderline, urine WBC 0-1, occasional bacteria). There is a mild rival cue (glucose 120, occasional bacteria), but these are clearly non-diagnostic. Diagnosis follows directly from the one distinguishing feature. This sits in band 2 territory---recognition plus mild exclusion work---perhaps low band 2 or high band 1. The lab table is mostly scenery reinforcing exclusion, i.e., redundant support for the key.
\end{description}

\subsection{Pairwise Difficulty Comparison}

\noindent USMLE items 103 and 392, judged by Nemotron 3 Super 120B. The empirical difficulty gap was 1.88 \textit{SD}, and item 392 was the harder item. Item 103 was presented as Question~A and item 392 as Question~B.
\begin{description}\small
\item[Unguided (chose item 392; correct).] Question B presents a patient with rheumatoid arthritis on immunosuppressive therapy who develops an acute, painful, warm elbow with fever and a synovial fluid leukocyte count of 60,000/mm³ but a negative Gram stain. The key difficulty lies in recognizing that this picture is most consistent with septic arthritis despite the negative Gram stain and the background of chronic inflammatory arthritis. Examinees must avoid being misled by the rheumatoid arthritis history into choosing an anti-inflammatory agent (colchicine, methylprednisolone, or NSAID) and instead recall that immunosuppressed patients with a monoarticular joint effusion and high fluid WBC require empiric antibiotics covering likely organisms (e.g., Staphylococcus aureus, including MRSA), making vancomycin the correct choice. This requires integrating immunosuppression risk, interpreting synovial fluid analysis, and knowing empiric antibiotic coverage---steps that are less straightforward than the immediate management in Question A. Question A, while also requiring recognition of heat-related illness, presents a more classic scenario: an adolescent with exertional hyperthermia, diaphoresis, and altered mental status. The management options are relatively limited and well-known (cooling, fluids, imaging, lumbar puncture, salt). Even if the correct answer is intravenous fluids (due to hypotension), the distractors are clearly less plausible, and the core concept---addressing the acute physiologic derangement---is more intuitive for most examinees. Consequently, fewer examinees are likely to be tripped up by the nuances in Question A compared with the diagnostic and therapeutic reasoning needed in Question B. Therefore, Question B is the harder item.
\item[Rubric-guided (chose item 392; correct).] Both questions contain an explicit fact in the stem that points to a specific wrong answer that the correct answer must override, which by itself tends to increase difficulty. In Question A the fact ``He vomited once after drinking water'' directly indicates that oral rehydration is not feasible, making intravenous fluids the clear next step and undermining the tempting choice of an ice-water bath. In Question B the fact that the patient is on chronic prednisone (and methotrexate) signals immunosuppression, raising the suspicion of septic arthritis despite a negative Gram stain and leading to the need for empiric antibiotics; this fact counters the intuitive choice of methylprednisolone. However, the difficulty of the two items diverges in the nature of the final step. Question A asks for a general therapeutic maneuver (intravenous fluids) that can be reasoned from the stem once the diagnosis of exertional heat stroke is recognized; the answer does not require recall of a specific drug name or regimen. Question B, by contrast, demands the selection of a particular antimicrobial agent (vancomycin) after diagnosing septic arthritis in an immunosuppressed patient. Retrieving a specific drug name is inherently more costly than choosing a broad intervention, and examinees are more likely to falter either by missing the infectious etiology in the setting of rheumatoid arthritis or by choosing an anti-inflammatory agent (e.g., methylprednisolone) instead of the correct antibiotic. Because Question B requires the retrieval of a specific drug name and involves a less direct clinical inference, it is the harder of the two items.
\end{description}

\section{Induced Absolute Judgment Rubrics}\label{app:absolute}

\subsection{USMLE Difficulty Rubric}

\bmhead{Overall characterization}
Across all five bands the surface features (stem length, presence of lab tables, patient age, specialty) do not track difficulty. What tracks difficulty is the relation between the most salient cue in the stem and the text of the keyed option: how many links separate them, whether an actively cued rival option exists, and whether a specific number or phrase in the stem argues against the key.

\subsubsection*{Band 1: Percentile 0--20}

\bmhead{Dominant pattern}
The stem contains a hallmark that maps onto the keyed option with no intervening link: the option text is essentially the name of the thing the hallmark denotes (mastoiditis for a protruding auricle after treated otitis media; neuroleptic malignant syndrome for antipsychotic plus fever, rigidity, elevated CK; common fibular nerve for foot drop after a blow to the fibular neck). Distractor options are drawn from other organ systems or other disease families, so that once the hallmark is recognized only one option is even in the right category. Quantitative items in this band are solved by a single arithmetic operation on a supplied 2×2 table (specificity, positive predictive value). Management items ask for the one canonical first action for the recognized entity (self-injectable epinephrine, thoracentesis for a large effusion, discontinue lisinopril for a new dry cough).

\bmhead{Exceptions}
About a quarter of the band consists of communication, ethics, and systems items in which no recognition occurs at all and the key is the professionally mandated action: certified/telephone interpreter rather than a family member, refusing to misrepresent examination findings on an insurance request, halting an operation until a laterality discrepancy is resolved, refusing to deceive an adolescent about a toxicology screen, ordering a medical evaluation for a colleague showing cognitive decline. These are placed here because the mandated action is stated in one option and every other option violates it.

\bmhead{Within-band variation}
How many independent findings in the stem redundantly point to the same option. At the low end four or five separate statements all converge on one answer and nothing in the stem is irrelevant; at the upper end a single sentence carries the whole answer and the remaining text is scenery or absent. This is a property internal to the band, not the finer grain of the band-1/band-2 boundary.

\bmhead{Representative items}
\medskip\noindent\textit{Item 66, percentile 0.} A deaf woman's hearing son telephones to arrange her appointment and offers to sign for her; the question asks how to ensure good communication.
\begin{itemize}\setlength{\itemsep}{0pt}\setlength{\parskip}{0pt}
\item[A.] Have the patient bring her son to the appointment to interpret
\item[B.] Arrange for a certified interpreter
\item[C.] Provide the patient with writing materials during the appointment
\item[D.] Talk with the patient face-to-face, enunciating carefully so that she may lip read
\item[E.] Obtain a history via a telecommunication device for the deaf before the appointment
\end{itemize}
\noindent Answer: B. Maximum redundancy---deafness since infancy, a third-party interpreter, and a sensitive gynecologic complaint each independently exclude every option but one, placing it at the floor of the band.

\medskip\noindent\textit{Item 94, percentile 11.} Substernal pain worsened by inspiration and relieved by leaning forward, fever, distant heart sounds, diffuse ST elevation in all leads, normal chest x-ray; which tissue is injured.
\begin{itemize}\setlength{\itemsep}{0pt}\setlength{\parskip}{0pt}
\item[A.] Aortic intima
\item[B.] Esophageal sphincter
\item[C.] Myocardium
\item[D.] Pericardium
\item[E.] Pleura
\end{itemize}
\noindent Answer: D. Three or four convergent cues, but the option is a tissue rather than the diagnosis, so redundancy is high yet slightly less complete than at the floor---mid-band.

\medskip\noindent\textit{Item 543, percentile 18.} Mice lacking a specific eosinophil protein harbor more adult filarial worms in lymph nodes than wild-type mice; what do they lack.
\begin{itemize}\setlength{\itemsep}{0pt}\setlength{\parskip}{0pt}
\item[A.] Cathelicidin-related antimicrobial peptide
\item[B.] Lysozyme
\item[C.] Major basic protein
\item[D.] Myeloperoxidase
\item[E.] $\alpha$-Defensin
\end{itemize}
\noindent Answer: C. A single cue (``protein found in eosinophils'') carries the entire item with no redundancy, which puts it at the ceiling of a band otherwise made of over-determined stems.

\subsubsection*{Band 2: Percentile 20--40}

\bmhead{Dominant pattern}
The diagnosis is as immediate as in band 1, but the question asks for one further link out from it: the mediator, the microbial property, the metabolite, the vessel, the anticipated complication, or the specific first drug. Gram-negative diplococci in a man with a third urethritis are handed to the examinee; what is asked is why reinfection occurs. Aspirin sensitivity in an asthmatic is transparent; what is asked is which mediator accumulates. Because the question moves one step past the label, the distractors are drawn from inside the same domain (other eicosanoids, other oral hypoglycemics, other neck arteries), so category matching no longer eliminates them. Also characteristic here are option sets presented as grids of arrows or normal/increased/decreased triads that must be assembled from three or four separate values.

\bmhead{Exceptions}
Roughly one item in six is a communication item in which two options are defensible and the key is the one that elicits the patient's own account before advice is given (``Tell me what you know about the steroid inhalers,'' ``what sorts of things were you doing at first,'' ``Are there any reasons why you might want to quit,'' exploring barriers to diet adherence). These sit here rather than in band 1 because the discarded options are also polite and clinically reasonable.

\bmhead{Within-band variation}
How many options remain live once the diagnosis has been fixed. At the low end the diagnosis knocks out three or four options because they belong to other diseases; in the middle two remain genuinely competitive; at the upper end every option is a real mediator, drug, or structure used in that same disease, so the diagnosis buys nothing. This is the band-1/band-2 boundary property at finer grain.

\bmhead{Representative items}
\medskip\noindent\textit{Item 59, percentile 21.} Sexually active man, purulent urethral discharge with intracellular gram-negative diplococci, third similar episode in 2 years; which organism property explains reinfection.
\begin{itemize}\setlength{\itemsep}{0pt}\setlength{\parskip}{0pt}
\item[A.] Antigenic variation
\item[B.] Catalase
\item[C.] Inhibition of B-lymphocyte function
\item[D.] Inhibition of T-lymphocyte function
\item[E.] Polysaccharide capsule
\end{itemize}
\noindent Answer: A. Naming the organism removes most options (catalase and capsule attach to other pathogens), so few remain live---low in the band.

\medskip\noindent\textit{Item 500, percentile 23.} Woman on metronidazole for trichomoniasis drinks champagne and vomits for 12 hours; also takes metformin, lisinopril, hydrochlorothiazide, acetaminophen.
\begin{itemize}\setlength{\itemsep}{0pt}\setlength{\parskip}{0pt}
\item[A.] Accumulation of acetaldehyde
\item[B.] Gastritis secondary to alcohol
\item[C.] Lactic acidosis secondary to metformin
\item[D.] Metronidazole-induced hepatotoxicity
\item[E.] Viral hepatitis
\end{itemize}
\noindent Answer: A. Two or three options (alcohol gastritis, metformin lactic acidosis) stay competitive because the stem supplies alcohol and metformin---mid-band.

\medskip\noindent\textit{Item 502, percentile 35.} Active ankylosing spondylitis unresponsive to several NSAIDs and to sulfasalazine; the next drug inhibits which target.
\begin{itemize}\setlength{\itemsep}{0pt}\setlength{\parskip}{0pt}
\item[A.] CD20
\item[B.] Cyclooxygenase-2
\item[C.] Cytotoxic T-lymphocyte antigen 4
\item[D.] Epidermal growth factor
\item[E.] Interleukin-1 (IL-1)
\item[F.] Tumor necrosis factor $\alpha$
\end{itemize}
\noindent Answer: F. Every option except EGF is a genuine rheumatologic target, so identifying the disease eliminates almost nothing---ceiling of the band.

\subsubsection*{Band 3: Percentile 40--60}

\bmhead{Dominant pattern}
No single finding carries the item; two or more separate stem elements (a drug plus a lab trend, an exposure plus a time course, a value plus a response to a maneuver) must be held together before any option can be excluded, and at least one option fits a subset of the data correctly. A platelet count alone, a graft thrombosis alone, or heparin alone would each point elsewhere; only the three together select the key. Correspondingly, keys are often stated as processes or categories rather than names---``decreased absorption,'' ``caspase activation,'' ``class I MHC recognized by CD8$^{+}$ lymphocytes,'' ``displacement of the nucleus pulposus''---so the examinee must convert an assembled clinical picture into an abstraction rather than into a diagnosis.

\bmhead{Exceptions}
Perhaps a fifth of the band is plain single-fact recall that lands here anyway because the fact is infrequently drilled: stool microscopy for a migrating perianal serpiginous rash with eosinophilia, Haemophilus ducreyi for a painful ulcer with satellite lesions, androgen insensitivity for absent sexual hair with normal breasts. These read like band-1 items and their placement is not explained by data assembly.

\bmhead{Within-band variation}
The number of unstated intermediate steps between the assembled stem and the wording of the key. One step at the low end (assemble the findings, read the diagnosis off the option list); two in the middle (assemble, name the mechanism, then classify it); three at the top (assemble, name the disorder, localize or reclassify it into a term the stem never uses). This is the band-2/band-3 boundary property at finer grain.

\bmhead{Representative items}
\medskip\noindent\textit{Item 203, percentile 40.} Two days after femoral-tibial bypass with a synthetic graft and continuous heparin, the graft has no pulse and the foot is ischemic; platelets have fallen from 250,000 to 60,000/mm³ with only mildly deranged coagulation times.
\begin{itemize}\setlength{\itemsep}{0pt}\setlength{\parskip}{0pt}
\item[A.] Disseminated intravascular coagulation
\item[B.] Factor IX deficiency
\item[C.] Heparin-induced thrombocytopenia
\item[D.] Idiopathic protein C deficiency
\item[E.] von Willebrand disease
\end{itemize}
\noindent Answer: C. One step after assembly---the diagnosis is spelled out in an option, so it sits at the band floor despite requiring three data points.

\medskip\noindent\textit{Item 450, percentile 46.} Stable hypothyroidism on levothyroxine becomes symptomatic with TSH 16.6 $\mu$U/mL after calcium carbonate was begun 2 months ago; which pharmacokinetic change explains it.
\begin{itemize}\setlength{\itemsep}{0pt}\setlength{\parskip}{0pt}
\item[A.] Decreased absorption
\item[B.] Decreased distribution
\item[C.] Decreased elimination
\item[D.] Increased distribution
\item[E.] Increased elimination
\item[F.] Increased metabolism
\end{itemize}
\noindent Answer: A. Two steps: link the new supplement to the failing drug, then translate binding in the gut into a pharmacokinetic category---mid-band.

\medskip\noindent\textit{Item 569, percentile 56.} One week of hourly urination and thirst; urine osmolality 50 mOsmol/kg that normalizes after ADH; which structure is dysfunctional.
\begin{itemize}\setlength{\itemsep}{0pt}\setlength{\parskip}{0pt}
\item[A.] Anterior pituitary gland
\item[B.] Bowman capsule
\item[C.] Glomerulus
\item[D.] Hypophysial portal system
\item[E.] Loop of Henle
\item[F.] Supraoptic nucleus
\end{itemize}
\noindent Answer: F. Three steps: read the water-deprivation logic, conclude central diabetes insipidus, then localize ADH synthesis when ``posterior pituitary'' is deliberately absent from the list---top of the band.

\subsubsection*{Band 4: Percentile 60--80}

\bmhead{Dominant pattern}
The keyed answer is still the canonical answer for the correctly identified problem, but the stem actively cues a rival option, and that rival appears in the list. A potassium of 7.3 with a creatinine of 1.8 cues renal failure while the key is drug effect; an adrenal mass with a raised aldosterone cues spironolactone while the key is alpha-blockade before resection; a mildly elevated intrapartum blood pressure and a hematocrit of 33 cue methyldopa and iron while the key is newborn-care education. A second, smaller subgroup asks for a granular anatomic or pharmacologic particular that is one level below the usual teaching point: which tendon externally rotates the shoulder, which trunk supplies the inferior parathyroid arteries, which tibial tubercle structure is avulsed, which vessel carries colonic metastases to the lung.

\bmhead{Exceptions}
A minority---roughly a fifth---are single-fact items whose content belongs in band 1 or 2 but whose keyed option is worded abstractly rather than nominally: ``epithelial tissue proliferation'' for cholesteatoma, ``decreased gluconeogenesis'' for galactosemia, ``transient lactase deficiency'' after enteritis, atrial septal defect from a fixed split S2. Nothing in these stems is a trap; the oblique option wording is what places them here.

\bmhead{Within-band variation}
How explicitly the stem foregrounds the rival option. Low in the band the rival is merely present in the list and mildly supported; in the middle a laboratory number in the stem directly names the rival's usual diagnosis; at the top the stem states, in words, the finding that constitutes the rival's own indication. This is the band-3/band-4 boundary property at finer grain; the band-4/band-5 boundary is different, because in band 4 the key remains the textbook action.

\bmhead{Representative items}
\medskip\noindent\textit{Item 42, percentile 62.} Sixteen-year-old on the morning after delivery; mildly elevated intrapartum blood pressure, 500 mL third-stage loss, BP now 132/84, normal reflexes, firm fundus, hematocrit 33\%, bonding well.
\begin{itemize}\setlength{\itemsep}{0pt}\setlength{\parskip}{0pt}
\item[A.] Begin oral ferrous sulfate
\item[B.] Begin oral methyldopa
\item[C.] Institute fundal massage
\item[D.] Order daily sitz baths
\item[E.] Provide education for well-baby care
\end{itemize}
\noindent Answer: E. The rivals (iron, methyldopa) are only weakly supported by numbers that are explicitly described as resolved or normal---low in the band.

\medskip\noindent\textit{Item 177, percentile 69.} Heart failure patient on lisinopril, spironolactone, digoxin, and metoprolol with weakness and K$^{+}$ 7.3 mEq/L, urea 40 mg/dL, creatinine 1.8 mg/dL, normal urinalysis.
\begin{itemize}\setlength{\itemsep}{0pt}\setlength{\parskip}{0pt}
\item[A.] Adverse effect of medications
\item[B.] Laboratory error
\item[C.] Metabolic acidosis
\item[D.] Renal failure
\item[E.] Rhabdomyolysis
\end{itemize}
\noindent Answer: A. Two numbers in the stem point toward the rival diagnosis, and a bicarbonate of 18 supports a second rival---mid-band.

\medskip\noindent\textit{Item 202, percentile 76.} Preoperative visit before adrenalectomy for a 10-cm mass; urinary normetanephrine/metanephrine and plasma aldosterone are elevated, BP 170/95, pulse 100/min.
\begin{itemize}\setlength{\itemsep}{0pt}\setlength{\parskip}{0pt}
\item[A.] Labetalol
\item[B.] A loading dose of potassium chloride
\item[C.] Nifedipine
\item[D.] Phenoxybenzamine
\item[E.] Spironolactone
\end{itemize}
\noindent Answer: D. The stem states the elevated aldosterone (spironolactone's indication) and the elevated pressure with tachycardia (labetalol's indication), so two rivals are named by the stem's own data---top of the band.

\subsubsection*{Band 5: Percentile 80--100}

\bmhead{Dominant pattern}
Here a specific datum or phrase in the stem argues against the keyed option, and the examinee must overrule it. A synovial leukocyte count of 20,000/mm³ with fever in a hospitalized patient must be read as gout; an ascitic neutrophil count only just above the diagnostic threshold must still be treated as spontaneous bacterial peritonitis; dysuria, dark urine, and fever in an elderly man on diuretics must be worked up as outlet obstruction with a postvoid residual rather than as infection; chronic disfiguring psoriasis with prior joint pain must be started on a topical agent; a chest x-ray with air-fluid level and an elevated stomach bubble but a midline mediastinum must be called non-tension. A related subgroup keys restraint or ``nothing further'': no imaging in uncomplicated pyelonephritis, no diagnostic study for a first urticaria, both nutrition strategies equally efficacious. Research-methods items in this band are conceptual cautions rather than computations---labeling a design (case series, open-label trial), invoking stratification for confounding, or rejecting a subgroup result because it was not the primary outcome.

\bmhead{Exceptions}
A substantial subset, perhaps a third, looks on the surface like band-1 material---one classic entity, five short options---and it belongs here because the stem withholds the cue students wait for: aortic dissection presented with a diastolic murmur and diminished femoral pulses but no tearing interscapular pain; subarachnoid hemorrhage with an entirely normal neurologic examination and no ``worst headache'' phrasing; cataplexy described only as legs collapsing while laughing; effort thrombosis in a jackhammer operator; transient tachypnea of the newborn with meconium-stained fluid and a treated maternal chlamydial infection in the stem. Their brevity and familiar-looking option lists do not lower their difficulty.

\bmhead{Within-band variation}
The strength with which a specific stated datum argues against the keyed answer. Low in the band, only the option list pulls elsewhere and the stem's numbers are neutral; in the middle, a borderline value in the stem appears to fall short of the threshold for the keyed action; at the top, a value or finding in the stem is unambiguously characteristic of the rejected option. This property operates inside band 5; the band-4/band-5 boundary is whether the key is the canonical action at all.

\bmhead{Representative items}
\medskip\noindent\textit{Item 18, percentile 81.} Two years of scaly, non-itchy red plaques on trunk, elbows, and knees with transient joint pain months ago, embarrassing at work, never treated.
\begin{itemize}\setlength{\itemsep}{0pt}\setlength{\parskip}{0pt}
\item[A.] Cyclosporine, orally
\item[B.] Methotrexate, orally
\item[C.] Phototherapy
\item[D.] Vitamin D, orally
\item[E.] Topical corticosteroids
\end{itemize}
\noindent Answer: E. Duration, extent, and prior arthralgia pull toward systemic therapy but no stated value contradicts a topical start---low in the band.

\medskip\noindent\textit{Item 637, percentile 87.} Cirrhotic man with fever to 37.8°C, diffuse abdominal tenderness, and ascitic fluid segmented neutrophil count of 288 cells/mm³; cultures pending.
\begin{itemize}\setlength{\itemsep}{0pt}\setlength{\parskip}{0pt}
\item[A.] Cefotaxime
\item[B.] Clindamycin
\item[C.] Metronidazole
\item[D.] Oxacillin
\item[E.] Vancomycin
\item[F.] No pharmacotherapy is indicated at this time
\end{itemize}
\noindent Answer: A. A number that reads as nearly normal, together with a near-normal temperature, argues for the ``no pharmacotherapy'' option---mid-band.

\medskip\noindent\textit{Item 172, percentile 99.} Hospitalized man with heart failure develops throbbing first-toe pain, erythema, and edema with a temperature of 38°C on day 2; joint aspirate analysis is asked for.
\begin{center}
\small
\begin{tabular}{@{}clll@{}}
\toprule
 & Leukocytes (/mm$^{3}$) & Crystals & Organisms \\
\midrule
A. & 100 & needle-shaped & none \\
B. & 100 & none & gram-positive cocci \\
C. & 100 & none & none \\
D. & 100 & rhomboid & none \\
E. & 20,000 & needle-shaped & none \\
F. & 20,000 & none & gram-positive cocci \\
\botrule
\end{tabular}
\end{center}
\noindent Answer: E. Fever plus a leukocyte count of 20,000/mm³ is the textbook signature of the rejected option (septic arthritis), and the key requires reading that same count as crystal-induced---top of the band.

\subsection{PANCE Difficulty Rubric}

\noindent\textit{Note.} The representative PANCE items are omitted from this version. All other rubric text is preserved.

\subsubsection*{Band 1: Percentile 0--20}

\bmhead{Dominant pattern}
Items here present a syndrome through a cluster of two or more findings that are individually near-pathognomonic and that co-occur only in the keyed condition, and the task is to name the condition, name its single classic association, or take the one step that is universally coupled to that condition. The stem supplies every element needed; nothing has to be inferred from an absent finding or from a normal value. The distractor set is drawn from categories that a first-pass reading already excludes---different organ systems, different acuity, or conditions that would have produced different named findings in the same stem. Where a drug, splint, or consult is asked for, exactly one option is the textbook-linked answer for the named condition and the others have no established link to it at all. Recall of a single fact-pair (condition $\rightarrow$ its classic feature, drug, or organism) is sufficient; no weighing of two plausible answers occurs.

\bmhead{Exceptions}
Roughly a fifth of the band is not diagnosis-recall but professional-conduct or health-policy items (fraud, implied consent, durable power of attorney, hand washing for hepatitis A). These share the property that only one option is legally or ethically defensible and the remainder are transparently improper, so they behave like the recall items despite different content.

\bmhead{Within-band variation}
The number of distinct stem findings that independently point at the keyed answer. At the low end, three or more redundant pointers appear (snoring + daytime somnolence + obesity question; upper-vs-lower-limb BP gap + murmur radiating to axilla + infant). Mid-band, two pointers do the work. At the upper end, a single pointer carries the item and the examinee must supply the rest of the syndrome from memory. This is the same property that separates Band 1 from Band 2, taken at finer grain---Band 2 items typically retain one pointer but require a second inference on top of it.

\subsubsection*{Band 2: Percentile 20--40}

\bmhead{Dominant pattern}
These items still name or point unambiguously at one condition, but the answer is one step removed from the condition name: a complication, a mechanism, a follow-on management decision, an adverse effect, or the one laboratory value that confirms rather than describes. The examinee must first identify the condition (usually easy) and then retrieve a second, less-rehearsed fact attached to it. Distractors are drawn from within the same organ system or the same drug class, so at least two options remain live after the diagnosis is made. Several items ask which of several correct-sounding statements about a condition is the one that actually holds---the discrimination is between adjacent facts, not between adjacent diseases.

\bmhead{Exceptions}
About a quarter of the band are one-step recall items whose surface looks like Band 1 (Zenker's location, aspirin inhibits platelet aggregation, hyponatremia in SIADH, cranial-nerve-free single-fact stems) yet sit here because the fact itself is less frequently drilled or because the correct option competes with a near-synonym in the list.

\bmhead{Within-band variation}
How far the keyed answer sits from the disease label along the chain presentation $\rightarrow$ diagnosis $\rightarrow$ attribute of the diagnosis. Low in the band, the answer is a defining attribute learned alongside the name (indirect hernia $\rightarrow$ incarceration; coxsackievirus $\rightarrow$ hand-foot-mouth vesicles). Mid-band, the answer is a management or drug choice that follows the diagnosis but has competing reasonable alternatives. High in the band, the answer is a second-order consequence or an epidemiologic association not stated in the stem (acanthosis nigricans as the physical sign of the metabolic picture; recheck-in-two-weeks rather than treat). This is a finer-grained version of the property that separates Band 2 from Band 1; it differs from what separates Band 2 from Band 3, which involves stems where the diagnosis itself is not settled by the stem.

\subsubsection*{Band 3: Percentile 40--60}

\bmhead{Dominant pattern}
Here the stem contains a finding or a history element whose significance must be decided before any option can be evaluated---the presentation is compatible with more than one of the listed answers until one detail is given weight. Typically two or three options are genuinely defensible on the presenting picture and the key is chosen by a single discriminating feature: a timing detail, a normal-versus-abnormal value, a prior procedure, or the fact that a finding has already resolved. Many items ask for the next step where several steps are all legitimate parts of the workup, so the answer turns on sequence rather than on correctness. Unlike Band 2, the disease label is often never settled at all; unlike Band 4, no option in the list is one most examinees would call wrong on principle.

\bmhead{Exceptions}
Perhaps a fifth of the band are pure single-fact recall items on lower-frequency content (spinal accessory nerve and shoulder shrug, salicylate ototoxicity, substantia nigra, Mendelian 25\% for two carriers) that carry no competition among options; they land here because the fact is less rehearsed, not because the reasoning is layered.

\bmhead{Within-band variation}
How many of the listed options remain defensible after the discriminating detail has been used. Low in the band, the detail eliminates all but one option cleanly and the residual competition is between the key and a single near-miss. Mid-band, two distractors survive the detail and must be ranked. High in the band, the key is a ``do nothing further / this is the physiologic answer'' option that competes against several active interventions, all of which are safe and reasonable. This property operates chiefly inside the band; the property distinguishing Band 3 from its neighbours is whether the diagnosis is settled at all by the stem.

\subsubsection*{Band 4: Percentile 60--80}

\bmhead{Dominant pattern}
Items here hinge on a fact that is not part of the standard presentation of the condition and cannot be reconstructed from the case: a specific administrative or regulatory number, a named surgical technique, an unexpected causal organism, a nerve level, a drug-class interaction, or a physiologic parameter used in a non-obvious direction. The clinical picture is frequently transparent, and the examinee still fails because the retrieved item is a detail attached to the condition only in specialised teaching. Options are homogeneous---five members of one class, five anatomic levels, five vaccines---so partial knowledge yields no elimination. Where Band 3 items reward correct weighting of stem detail, these reward possession of one discrete stored fact; where Band 5 items require holding a hidden second condition, these are visible on their face and simply unfamiliar.

\bmhead{Exceptions}
Roughly a quarter of the band are visual/waveform items and management items (chest x-ray tube thoracostomy, ECG premature atrial contractions, x-ray of proximal tibia fracture) that fail for a different reason---the finding must be read off an image and its significance stated, with no verbal cue available.

\bmhead{Within-band variation}
How narrowly the required fact is taught relative to the frequency of the condition it belongs to. Low in the band the fact is peripheral but rehearsed (Boutonnière description; rheumatoid deformity naming). Mid-band the fact is a mechanistic or classificatory detail attached to a common condition (mechanism of shock in sepsis; nerve level for a common radiculopathy). High in the band the fact is a numeric threshold, a legal source, or an association essentially disconnected from clinical reasoning (six months for hospice; state law for prescriptive authority; Wilson disease as the copper-IUD contraindication). This is the same axis that separates Band 4 from Band 3, at finer grain.

\subsubsection*{Band 5: Percentile 80--100}

\bmhead{Dominant pattern}
These stems contain a deliberate misdirection: the most salient element of the presentation is not the one the answer depends on, and the keyed option requires either overriding an alarming finding or attaching the case to a second, unmentioned condition. Normal or near-normal data are frequently supplied so that the correct response is to do nothing, to proceed, or to name the physiologic/benign explanation while active-sounding interventions occupy the option list. Several items ask which of a list of stated features is the risk factor or cause, where every listed feature is present in the patient and only one carries the association. The distinguishing move from Band 4 is that the required knowledge is usually ordinary; what fails is the decision about which stem element governs.

\bmhead{Exceptions}
About a fifth of the band are genuinely obscure single-fact items with no misdirection (byssinosis/asbestosis discrimination, C-reactive protein as the infectious marker in atherosclerosis, vitamin A toxicity with papilledema, actinic keratitis after skiing), which sit here purely because the association is rarely encountered.

\bmhead{Within-band variation}
The strength of the pull exerted by the discarded stem element relative to the governing one. Low in the band, the misleading element is mild and the governing element is at least mentioned (ambiguous genitalia in a neonate; nontender enlarging neck node referred rather than treated). Mid-band, the stem foregrounds a plausible diagnosis that must be set aside for a second one hidden in the history (heart murmur consultation that resolves into sleep apnoea; chronic steroid history that resolves into adrenal insufficiency). At the top, the keyed answer contradicts what the presenting emergency appears to demand or requires selecting a risk factor from a list in which several textbook risk factors are all present in the patient. This property operates only inside this band; what separates Band 5 from Band 4 is the presence of misdirection at all.

\section{Induced Pairwise Comparison Rubrics}\label{app:pairwise}

\subsection{USMLE}

You must choose one of the two items as harder. Work through each item alone first, then compare.

For each item on its own:

\noindent\textbf{1.} What the final step actually asks for: a diagnosis, a mechanism, a specific drug or regimen, which test comes first, a number, a design name, a statement to a patient.

\noindent\textbf{2.} Whether you could produce that answer before reading the options.

\noindent\textbf{3.} Whether any single sentence in the vignette names a fact---an exposure, an occupation, a recent drug, a meal, a prior infection, a laboratory value, a comorbidity---that points at one specific wrong option.

The strongest thing to look for. An explicitly stated fact in the stem that supports a specific named wrong answer which the correct answer must override. This is a real sentence you can quote, not a habit you attribute to examinees. Where you find one, that item is likely the harder of the two. ``Examinees are trained to image,'' ``people will want to do something,'' ``the answer is counterintuitive'' are not this and must not be used.

If the stem states the discriminating finding in plain words, or restates the answer in the stem, the diagnostic step costs nothing---then look at what is left and price that.

Price the final step on its own; never let an easy diagnosis discount it.

\begin{itemize}\setlength{\itemsep}{0pt}\setlength{\parskip}{0pt}
\item Retrieving a specific drug, route, or combination regimen for an already-obvious diagnosis is expensive. Treat ``the diagnosis is clear, so the therapy follows'' as a mistake. This applies to antimicrobial regimens, antidotes and prophylaxis, immunomodulators, growth factors, diuretic pairings, and symptom-relief-only answers.
\item Options that are near neighbours in one family are expensive: two muscles of the same group, two disorders of the same organ in the newborn, two sleep disorders, topical versus systemic form of the same drug class, two segments of one tube, two near-synonymous phrasings of a mechanism.
\item The reverse is cheap, however obscure the fact: one option in the right organ system, vessel group, or drug class with the rest plainly elsewhere.
\item Which test to obtain first, when imaging is premature, and what value triggers treatment are expensive.
\item ``Which of these true risk factors is greatest'' is expensive, unless only one option is actually a cause of the condition, in which case it collapses.
\end{itemize}
Communication, counselling and professionalism. If the answer is the open-ended question, the empathic reply, telling the patient the truth, bringing the patient into the decision, eliciting the patient's own barriers or own reasons, or arranging a supportive service, the item is cheap---however many other options look professionally reasonable. If the answer turns on a rule (who may lawfully receive information, a reporting duty, an exception to consent, testing a minor) or on end-of-life drug choice, it is expensive.

Restraint answers. Observation, reassurance, no further study, conservative measures only: cheap when the examination and laboratory results are unremarkable and the only abnormality is the patient's own worry. Expensive when you must override objective abnormalities---fever, leukocytosis, purulent findings, an abnormal image or tracing.

Statistics. Reading a conclusion off a printed table, interpreting a p-value, or reading confidence intervals for significance is cheap. Naming a study design, explaining what a subgroup finding does not establish, or constructing an interval around a mean where the work is choosing between the spread of individuals and the precision of the mean is expensive.

Stop treating these as decisive. Stem length. Number of options. Arrow, matrix or column layouts. A photograph or tracing that only illustrates what the text already states---a picture counts only if, with the text alone, you could not answer. And above all, do not treat your own instant recognition of the disease as evidence that examinees found it easy: many of the items they miss most often are the ones that read as textbook.

Apply the same basis to both items. If a consideration would change simply because the pair was presented in the opposite order, do not let it decide the comparison.

If the two still look level, choose the one whose answer had to be retrieved as a specific name---a drug, a regimen, a design, a named entity---over the one whose answer you could reason out from the stem. If they are still level, choose the one whose options are near neighbours of each other.

\subsection{PANCE}

You must choose one of the two items as harder. Work through the following for both items before deciding.

\noindent\textbf{1.} Ignore these entirely. They caused most of the wrong calls and can be written about either item.

\begin{itemize}\setlength{\itemsep}{0pt}\setlength{\parskip}{0pt}
\item Length of the stem, amount of laboratory data, number of reasoning steps.
\item Presence of a photograph, radiograph or tracing.
\item Whether the question asks for a diagnosis, a next step, or a treatment.
\item Number of options.
\item ``There is one genuinely tempting distractor''---true of nearly every item.
\item ``The other options are absurd, so it can be answered by elimination''---items with implausible distractors were often the harder one.
\item ``The keyed answer differs from what clinicians actually do.''
\end{itemize}
\noindent\textbf{2.} Place in the upper half by default (harder):

\begin{itemize}\setlength{\itemsep}{0pt}\setlength{\parskip}{0pt}
\item Questions with no patient, or only a one-line patient, that ask for a mechanism of drug action, the pathophysiologic mechanism of a disease or a shock state, what a hormone or system responds to, a definition, a diagnostic criterion (``which finding indicates / is most characteristic of / is most specific for''), a category label (an isolation category, a complication name, an ethical or legal principle), or a fixed number or eligibility threshold. Do not downgrade these because the fact feels elementary to you.
\item Items whose entire content is matching a described finding to its name, or choosing the right term among adjacent terms from one vocabulary family---surgical complications, dementia subtypes, shock categories, anxiety disorders, renal syndromes, ethical principles. Reciting the textbook descriptor inside the stem does not make such an item easy.
\item Items where two or more options are genuinely true of this patient and the keyed one wins only by degree (two drug classes that both raise a level, two adverse effects that both occur, two real risk factors).
\item Items where the question quietly shifts target---prophylaxis for contacts rather than treatment of the patient, the underlying condition rather than the visible lesion, a benign-looking complaint whose keyed answer is an unstated malignancy.
\item Stems asking which drug is contraindicated, which would worsen the situation, or which finding argues against the diagnosis.
\item Answers that require acting against a reflex learned for an adjacent condition---but only if the stem does not itself spell out the conflicting fact. If the stem lists the contraindication or the disqualifying history, the override has been handed over and the item is easy.
\end{itemize}
\noindent\textbf{3.} Place in the lower half by default (easier):

\begin{itemize}\setlength{\itemsep}{0pt}\setlength{\parskip}{0pt}
\item A patient stem containing one distinctive finding that only the keyed answer explains---a specific auscultatory event, a striking quantitative laboratory pattern, a characteristic exposure or timing---where the question asks for the diagnosis, the drug, or the action.
\item Selecting a test or a treatment inside a full vignette whose diagnosis is unmistakable and whose standard test or first-line drug is well drilled, even when a gold-standard, more invasive, or later-in-workup option is on the list. That structure is not a trap here. (Test-selection becomes mid-to-hard only when the stem is a one-liner or names a condition with no clinical picture.)
\item An emergency with one obvious action.
\item Deterministic interpretation with a fixed rule (acid--base, simple rhythm or film patterns that are grossly abnormal).
\item An unambiguously wrong professional act with an obvious refusal.
\end{itemize}
\noindent\textbf{4.} For two diagnosis items, use the option list against the vignette. Count how many listed diagnoses could occur in this patient, of this age, with this headline complaint. If two or more could, and separating them depends on tempo, on an absence, or on a qualitative adjective, treat the item as hard---a memorised triad or a classic association does not rescue it. If one stem finding is explained by the keyed answer alone, treat it as easy. A detail in the stem that argues against the keyed answer pushes it harder still.

\noindent\textbf{5.} Note that a pathognomonic cue only helps if the topic is one everyone drills. A one-to-one sign on a peripheral topic behaves like an unfamiliar label, not like an easy cue.

\noindent\textbf{6.} Weak signals, for breaking a tie only, never decisive on their own: the answer is observation or reassurance; the answer is a superlative among several real risk factors; how far the presentation departs from the most-illustrated version of that disease.

\noindent\textbf{7.} Apply the same basis to both items. Do not let stem length, number of reasoning steps, a supposedly strong distractor, image interpretation, or management-versus-recall framing decide the pair; these can be written about either item. Decide on points 2--4 instead.

\end{appendices}

\end{document}